\documentclass[letterpaper]{article} 
\usepackage{aaai23}  
\usepackage{times}  
\usepackage{helvet}  
\usepackage{courier}  
\usepackage[hyphens]{url}  
\usepackage{graphicx} 
\usepackage{natbib}  
\usepackage{caption} 
\usepackage{algorithm}
\usepackage{algorithmic}

\usepackage{pifont}
\usepackage{booktabs}
\usepackage{amsmath}

\usepackage{newfloat}
\usepackage{listings}
\DeclareCaptionStyle{ruled}{labelfont=normalfont,labelsep=colon,strut=off} 
\floatstyle{ruled}
\newfloat{listing}{tb}{lst}{}
\floatname{listing}{Listing}
\title{Learning Progressive Modality-shared Transformers for Effective Visible-Infrared Person Re-identification}
\author{
    Hu Lu\textsuperscript{\rm 1},
    Xuezhang Zou\textsuperscript{\rm 1},
    Pingping Zhang\textsuperscript{\rm 2}\thanks{The corresponding author.}
}
\affiliations{
    \textsuperscript{\rm 1}School of Computer Science and Communication Engineering, Jiangsu University\\
    \textsuperscript{\rm 2}School of Artificial Intelligence, Dalian University of Technology\\
    luhu@ujs.edu.cn;2222108016@stmail.ujs.edu.cn;zhpp@dlut.edu.cn
}

\usepackage{bibentry}

\begin{document}

\maketitle

\begin{abstract}
Visible-Infrared Person Re-Identification (VI-ReID) is a challenging retrieval task under complex modality changes.
Existing methods usually focus on extracting discriminative visual features while ignoring the reliability and commonality of visual features between different modalities.
In this paper, we propose a novel deep learning framework named Progressive Modality-shared Transformer (PMT) for effective VI-ReID.
To reduce the negative effect of modality gaps, we first take the gray-scale images as an auxiliary modality and propose a progressive learning strategy.
Then, we propose a Modality-Shared Enhancement Loss (MSEL) to guide the model to explore more reliable identity information from modality-shared features.
Finally, to cope with the problem of large intra-class differences and small inter-class differences, we propose a Discriminative Center Loss (DCL) combined with the MSEL to further improve the discrimination of reliable features.
Extensive experiments on SYSU-MM01 and RegDB datasets show that our proposed framework performs better than most state-of-the-art methods.
For model reproduction, we release the source code at https://github.com/hulu88/PMT.
\end{abstract}

\section{Introduction}
\begin{figure}[t]
\centering
\includegraphics[width=1.0\columnwidth]{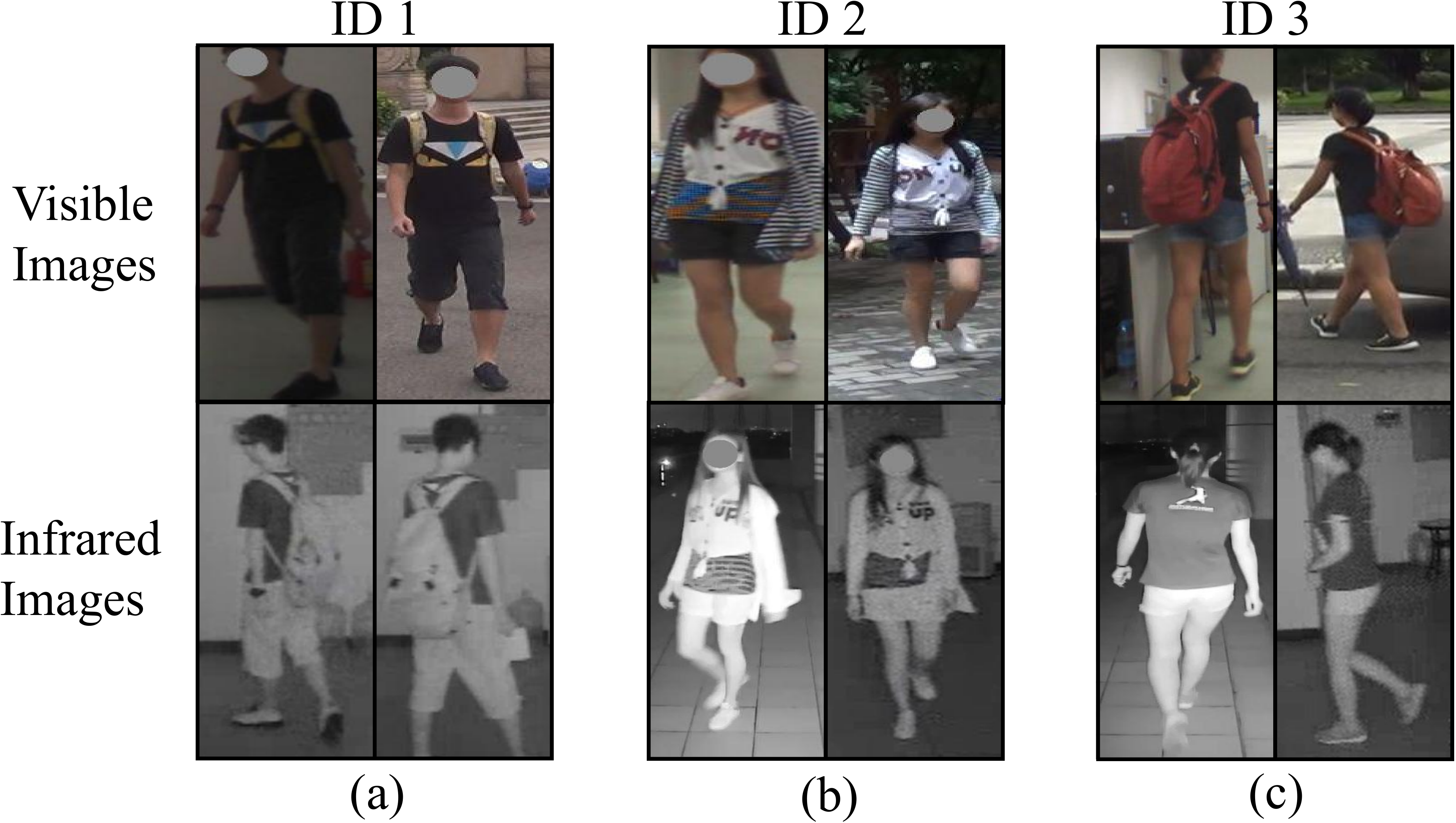}
\caption{Several typical cases in visible-infrared person re-identification.
	(a) Discriminative information is not always visible due to posture or viewpoint changes.
	(b) Partial information may disappear due to the modality shift and different lighting conditions.
	(c) Modality-based clothing change due to the large time span.
}
\label{fig:challenges}
\end{figure}
Person Re-Identification (ReID) aims to retrieve the same person under different cameras and times.
It can be utilized in many real-world applications, such as video surveillance, smart security, etc.
Recently, with the advances of deep learning, person ReID has witnessed great success in performance and deployment.
However, most of the existing ReID methods target on the visible environment.
Thus, they can be regarded as visible-visible ReID.
In fact, most of visible-visible ReID methods can not work well at nighttime.
To address this problem, images captured by infrared cameras are considered in practical scenarios, which greatly help the ReID under different modalities and result in Visible-Infrared Person Re-Identification (VI-ReID).

Compared with single-modality ReID, VI-ReID has three main challenges:
1) The large modality gap will make it difficult to align the identify-related features of the two modalities.
2) Infrared images are more sensitive to light conditions than visible images, resulting in less discriminative features for cross-modality matching.
3) Modality-based clothing changes can occur due to the large time span, which further increases the difficulty of robust feature extraction.

To reduce the heterogeneous differences between two modalities, existing approaches~\cite{ye2021deep,gao2021mso,chen2021neural} mainly use a dual-stream network structure.
The non-shared weight components are first used to extract modality-specific features separately before learning modality-shared features.
Although these methods can effectively benefit from modality-specific features and deal with inter-modality differences, they can hardly extract effective modality-shared features.
Meanwhile, there are also some Generative Adversarial Network (GAN)-based methods~\cite{li2020infrared,dai2018cross,wang2019rgb} that generate cross-modality images by learning modality transformed patterns.
However, these methods usually introduce additional image noises and huge computational costs.
Thus, they are difficult to deploy in practical scenarios.

In addition, some outstanding methods aim to extract more discriminative information.
For example, \cite{zhu2020hetero,zhang2021global} horizontally divide the portrait into multiple regions to align independent local features and focus on extracting fine-grained discriminative features.
However, due to the great challenges, overreliance on these features may lead to wrong matches, as shown in Fig.~\ref{fig:challenges}.
Therefore, recent VI-ReID works mainly focus on reducing the feature differences between modalities.

In this work, we propose a novel deep learning framework named Progressive Modality-shared Transformer (PMT) to extract reliable modality-invariant features for effective VI-ReID.
To this end, we first propose a progressive learning strategy with Transformers~\cite{dosovitskiy2020image} to reduce the gap between visible and infrared modalities.
More specifically, we improve the hard triplet loss and introduce gray-scale images as an auxiliary modality to learn modality-independent patterns.
Besides, we propose a Modality-Shared Enhancement Loss (MSEL) to reduce the negative effects of modality differences and enhance the features with modality-shared information.
Finally, we propose a Discriminative Center Loss (DCL) to deal with the large intra-class variance, further enhancing the discrimination of reliable modality-shared features.
Extensive experiments on SYSU-MM01 and RegDB datasets show that our framework performs better than most state-of-the-art methods.

Our main contributions are summarized as follows:
\begin{itemize}
\item We propose a novel deep learning framework (\emph{i.e., \textbf{PMT}}) for effective VI-ReID, focusing on extracting more robust modality-shared features.
\item We propose a new Modality-Shared Enhancement Loss (MSEL) to enhance the modality-shared features, thus effectively addressing the problem of feature unreliability.
\item We propose a new Discriminative Center Loss (DCL) to deal with large intra-class differences and further enhance the discrimination of modality-invariant features.
\item Extensive experimental results on the SYSU-MM01 and RegDB datasets show that our proposed method achieves a new state-of-the-art performance.
\end{itemize}
\section{Related Work}
\subsection{Visible-infrared Person ReID}
Visible-infrared person ReID aims to retrieve the same person under different image modalities.
In fact, Wu \emph{et al.}~\cite{wu2017rgb} first explicitly defined the VI-ReID task and built a large-scale and challenging dataset.
Existing VI-ReID methods usually adopt dual-stream networks and mine modality-shared features.
For example, Ye \emph{et al.}~\cite{ye2018hierarchical} propose an effective dual-stream network to explore modality-specific and modality-shared features simultaneously.
Lu \emph{et al.}~\cite{lu2020cross} propose a mechanism of feature information complementarity to exploit the potential of modality-specific features.
Gao \emph{et al.}~\cite{gao2021mso} propose a multi-feature space joint optimization network to enhance modality-shared features.
Zhang \emph{et al.}~\cite{zhang2021global} introduce a dual-stream network to achieve the global-local multiple granularity learning.
Based on dual-stream networks, Zhu \emph{et al.}~\cite{zhu2020hetero} propose a Heterogeneous-Center (HC) loss to reduce the modality gaps.
Liu \emph{et al.}~\cite{liu2020parameter} further design a heterogeneous-center triplet loss and explore the parameter sharing methods to improve the feature representation ability.
Ye \emph{et al.}~\cite{ye2020visible} introduce gray-scale images as auxiliary modalities and realize the homogeneous augmented tri-modal learning.
Fu \emph{et al.}~\cite{fu2021cm} propose the cross-modality neural architecture search and improve the structural effectiveness for VI-ReID.
Hao \emph{et al.}~\cite{hao2021cross} introduce a modality confusion mechanism and a center aggregation method to reduce the differences between modalities.
Meanwhile, many image generation-based methods are developed to mitigate the large modality gap.
For example, Dai \emph{et al.}~\cite{dai2018cross} introduce the GAN framework for cross-modality image generation, and propose the so-called cmGAN for feature learning.
Furthermore, Wang \emph{et al.}~\cite{wang2019rgb} propose the AlignGAN and convert visible images to infrared images with joint pixel and feature alignment.
Choi \emph{et al.}~\cite{choi2020hi} attempt to disentangle cross-modality representations with hierarchical structures.
Li \emph{et al.}~\cite{li2020infrared} introduce an auxiliary X-modality to generate robust features and bridge the different modalities.
Although the above methods are effective, they usually introduce additional image noises and huge computational costs.
Thus, they are difficult to deploy in practical scenarios.
\begin{figure*}[t]
\centering
\includegraphics[width=2.0\columnwidth]{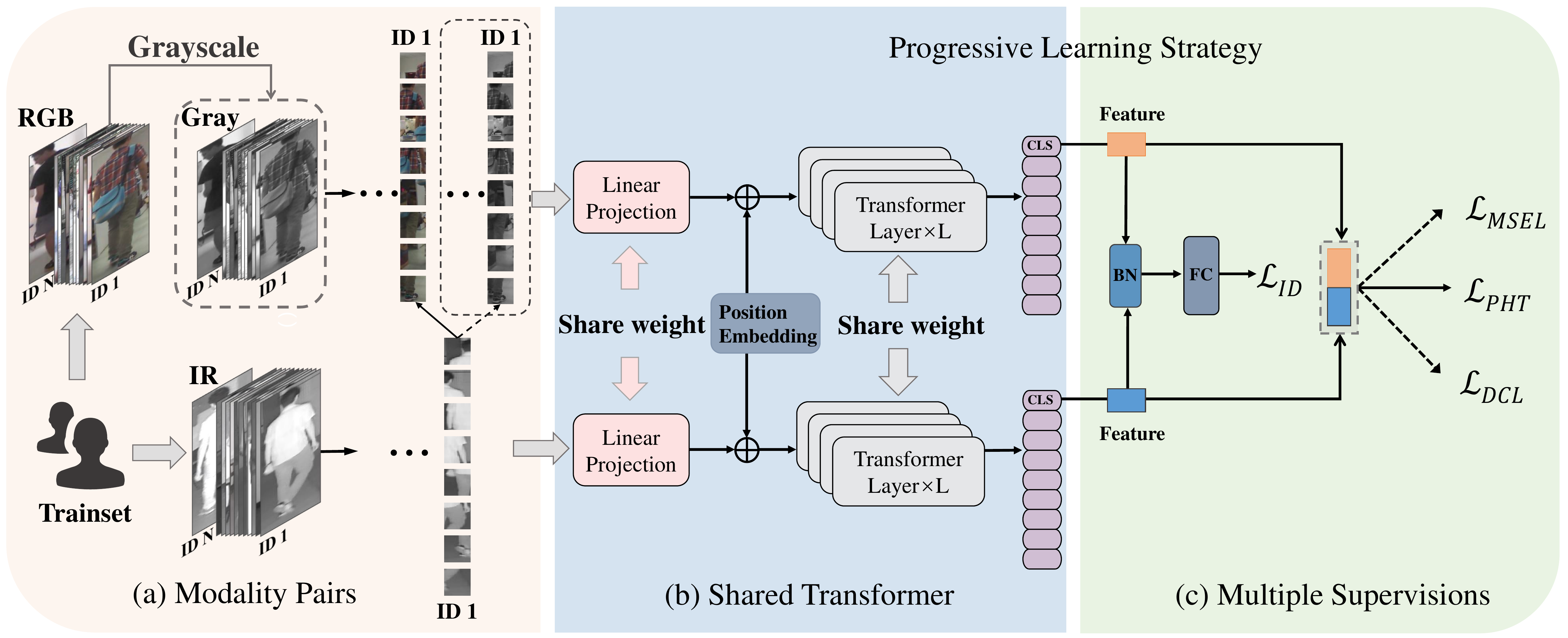}
\caption{
The framework of our proposed Progressive Modality-shared Transformer (PMT). To deal with the large modality gap, we propose a progressive learning strategy: 1) At the first stage, we feed gray-scale images and infrared images into a weight-shared Transformer supervised by $L_{ID}$ and $L_{PHT}$ for modality-independent feature extraction. 2) At the second stage, we utilize visual images and infrared images to improve the modality-shared features with $L_{MSEL}$ and $L_{DCL}$.
}
\label{fig:framework}
\end{figure*}
\subsection{Transformer in Person ReID}
Transformers~\cite{vaswani2017attention} are initially proposed in Natural Language Processing (NLP).
Recently, they have been utilized for some computer vision tasks, including person Re-ID.
For visible-visible person ReID, He \emph{et al.}~\cite{he2021transreid} improve the Vision Transformer (ViT)~\cite{dosovitskiy2020image} with a side information embedding and a jigsaw patch module to learn discriminative features.
Zhu \emph{et al.}~\cite{zhu2021aaformer} add the learnable vectors of part tokens to learn part features and integrate the part alignment into the self-attention.
Lai \emph{et al.}~\cite{lai2021transformer} utilize Transformers to generate adaptive part divisions.
%
%
Zhang \emph{et al.}~\cite{zhang2021hat} propose a hierarchical aggregation Transformer for integrating multi-level features.
%
%
Chen \emph{et al.}~\cite{chen2021oh} propose an omni-relational high-order Transformer for person Re-ID.
Ma \emph{et al.}~\cite{ma2021pose} propose a pose-guided inter-and intra-part relational Transformer for occluded person Re-ID.
As for VI-ReID, Liang \emph{et al.}~\cite{liang2021cmtr} improve the single-modality Transformer and try to remove modality-specific information.
Chen \emph{et al.}~\cite{chen2022structure} utilize the human key-point information and introduce a structure-aware positional Transformer to learn semantic-aware modality-shared features.
Although all the above Transformer-based methods have achieved superior performances, they generally lack of desirable modality-invariant properties.
In this work, we explore the progressive modality-shared Transformers and learn reliable features for the VI-ReID task.
\section{The Proposed Method}
In this section, we introduce the details of the proposed Progressive Modality-shared Transformer (PMT) for VI-ReID.
As shown in Fig. \ref{fig:framework}, we first take the gray-scale images as an auxiliary modality and adopt a weight-shared ViT~\cite{dosovitskiy2020image} as our feature extractor to capture modality-invariant features.
Then, we propose a progressive learning strategy to deal with the large modality gap.
Besides, a Modality-Shared Enhancement Loss (MSEL) is employed for enhancing modality-shared features.
Finally, a Discriminative Center Loss (DCL) is introduced to further improve the discrimination of reliable modality-shared features.
\subsection{Progressive Learning Strategy}
Although previous weight-shared structures can capture more modality-shared features, they are also susceptible to modality-specific noises.
Besides, the pre-trained weights on ImageNet~\cite{deng2009imagenet} generally have a stronger reliance on low-level features, such as color or texture.
%
Thus, directly using these pre-trained models may miss some modality-specific information.
Considering above facts, we design a progressive learning strategy.
The key idea is to remove color information of visible images through gray-scale images.
It also helps to learn the modality-independent discriminative patterns.
In this way, the negative effects from large modality gaps are effectively mitigated.

Formally, we denote the visible image and the infrared image as $x^{vis}$ and $x^{ir}$, respectively.
Then, the gray-scale image corresponding to $x^{vis}$ can be denoted as $x^{gray}$.
By feeding \{$x^{vis}$, $x^{gray}$, $x^{ir}$\} into a weight-shared Transformer $\mathcal{F(\cdot)}$, we can obtain their corresponding embedding vectors:
\begin{equation}
	f^{v}=\mathcal{F}\left(x^{vis}\right), f^{g}=\mathcal{F}\left(x^{gray}\right),f^{i r}=\mathcal{F}\left(x^{ir}\right).
\end{equation}
To reduce the impact of modality-specific information, we further propose a Progressive Hard Triplet Loss (PHT).
Similar to most VI-ReID methods, in each mini-batch, we randomly select $P$ identities and then select each identity's $K$ visible and $K$ infrared images.
Then, the proposed progressive hard triplet loss can be denoted as:
\begin{equation}
	L_{PHT}=\left\{\begin{array}{ll}
		L_{Intra}, & X=\left\{x^{gray}, x^{ir}\right\}\\
		L_{Global}, & X=\left\{x^{vis}, x^{ir}\right\}
	\end{array}\right.
\end{equation}	
\begin{equation}
	\begin{aligned}
		{L_{Intra}}= \sum_{i=1}^{PK} \left[\max _{\forall y_i=y_j} D\left(f_{i}^{g}, f_{j}^{g}\right)-\min_{\forall y_i \neq y_k} D\left(f_{i}^{g}, f_{k}^{g}\right)+m\right]_{+} \\
		+\sum_{i=1}^{PK} \left[\max _{\forall y_i=y_j} D\left(f_{i}^{ir}, f_{j}^{ir}\right)-\min _{\forall y_i \neq y_k} D\left(f_{i}^{i r}, f_{k}^{i r}\right)+m\right]_{+}.
	\end{aligned}
\end{equation}
\begin{equation}
	{L}_{Global}=\sum_{i=1}^{2PK} \left[\max _{\forall y_i=y_j} D\left(f_{i}, f_{j}\right)-\min _{\forall y_i \neq y_k} D\left(f_{i}, f_{k}\right)+m\right]_{+}.
\end{equation}
where $D(\cdot, \cdot)$ represents a distance metric.
%
$y_i$ is the identity label of the $i$-th image.
$[z]_{+}=\max(z, 0)$.
$m$ is a margin.
\begin{figure}[t]
	\centering
	\includegraphics[width=1.0\columnwidth,trim=0 0 10 10]{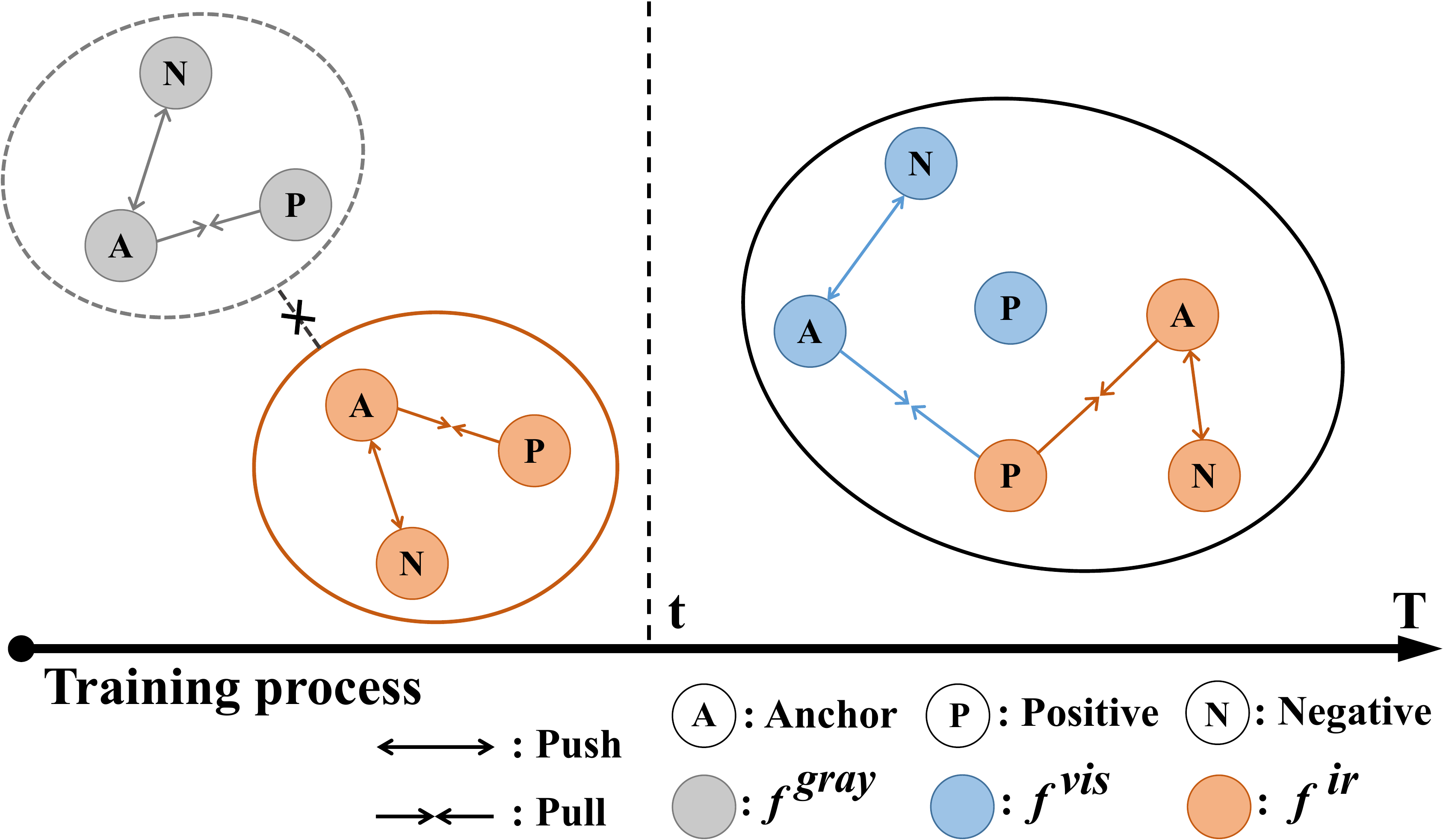}
	\caption{Illustration of the Progressive Learning Strategy.}
	\label{fig:pls}
\end{figure}

As shown in Fig.~\ref{fig:pls}, we divide the entire training process into two stages.
At the first stage, the framework takes \{$x^{gray}$, $x^{ir}$\} as input, independently sampling positive and negative samples within each modality.
With $L_{intra}$, the framework mainly focuses on learning the modality-independent discriminative patterns, thus effectively alleviating the negative effects caused by the large gap between visible and infrared modalities.
At the second stage, we replace the inputs with \{$x^{vis}$, $x^{ir}$\} to take full advantages of modality-specific information for more fine-grained learning.
With $L_{global}$, the framework will no longer distinguish different modalities and select positive and negative samples only based on feature distances.
This can keep the raw image information, and allow the model to be benefited from modality-specific information.
\begin{figure}[ht]
	\centering
	\includegraphics[width=1.0\columnwidth,trim=0 0 10 10]{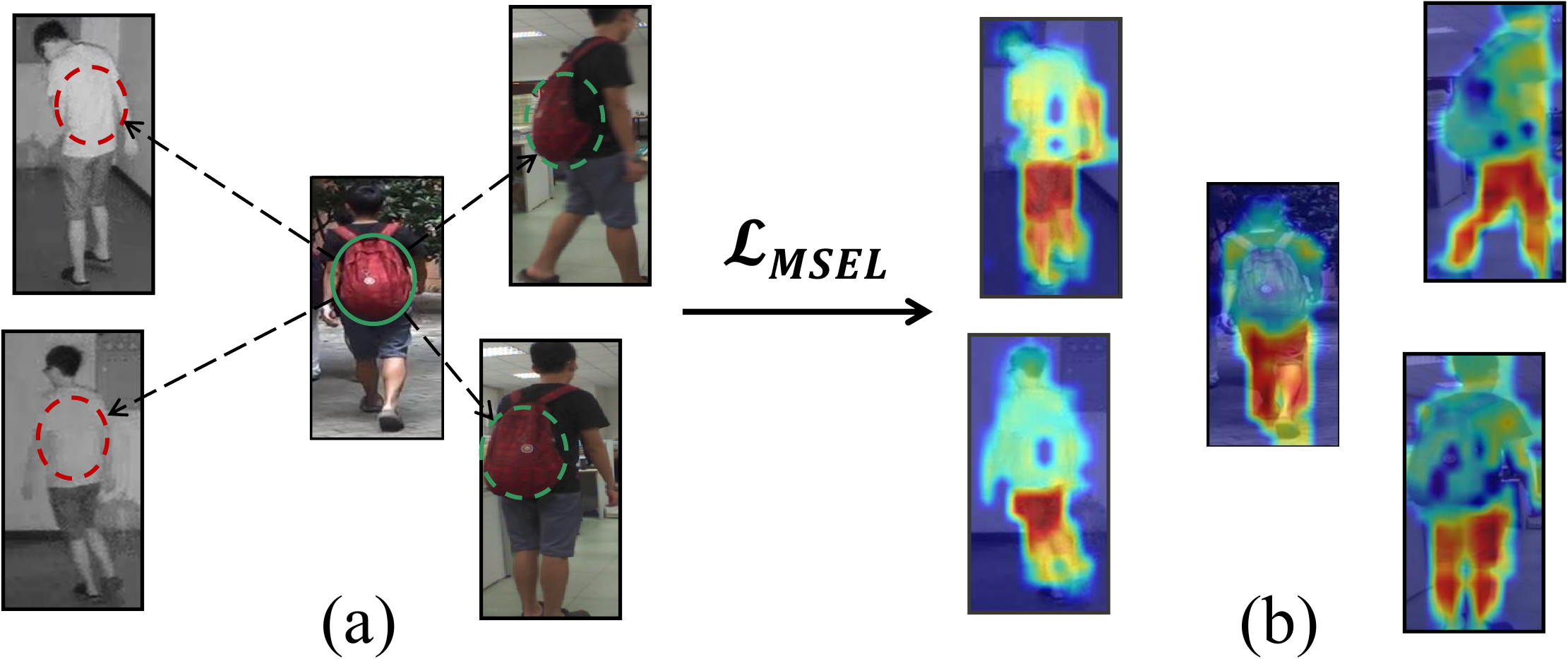}
	\caption{Motivations of the proposed Modality-Shared Enhancement Loss. It suppresses the unreliable features that only appear in one modality and enhances the modality-shared features.}
	\label{fig:msel}
\end{figure}
\subsection{Modality-Shared Enhancement Loss}
In real scenes, there are large modality differences between visible and infrared images.
Thus, it is essential to extract modality-invariant features.
%
%
As shown in Fig.~\ref{fig:msel} (a), the red backpack appears only in the visible modality, so over-reliance on such features will lead to failure in cross-modality retrieval.
%
%
Therefore, we introduce the MSEL to appropriately suppress the unreliable features that only appear in one modality and enhance the utilization of reliable modality-invariant features.

To the above goal, we explore potential information from all samples in a mini batch.
Formally, we denote the anchor features of the infrared and visible modality as $f^{ir}_{a}$ and $f^{vis}_{a}$, respectively.
Without loss of generality, we take $f^{ir}_{a}$ as an example.
Firstly, we calculate its average distance to other positive samples under the intra modality and cross modality, denoted as:
\begin{equation}\label{eqintra}
\mathrm{D}^{intra}=\frac{1}{K-1} \sum_{\substack{i=1 \\ i \neq a}}^{K} D\left(f_{a}^{ir}, f_{i}^{ir}\right),
\end{equation}

\begin{equation}\label{eqcross}
\mathrm{D}^{cross}=\frac{1}{K} \sum_{i=1}^{K} D\left(f_{a}^{ir}, f_{i}^{vis}\right).
\end{equation}
%
Then, the $L_{MSEL}$ is defined as:
\begin{equation}\label{eqmsel}
L_{MSEL}=\frac{1}{2PK} 	\sum_{p=1}^{P}\left[\sum_{a=1}^{2K}\left(\mathrm{D}^{intra}_a-\mathrm{D}^{cross}_a\right)^{2}\right].
\end{equation}

\begin{figure}[t]
\centering
\includegraphics[width=1.0\columnwidth,trim=0 0 10 10]{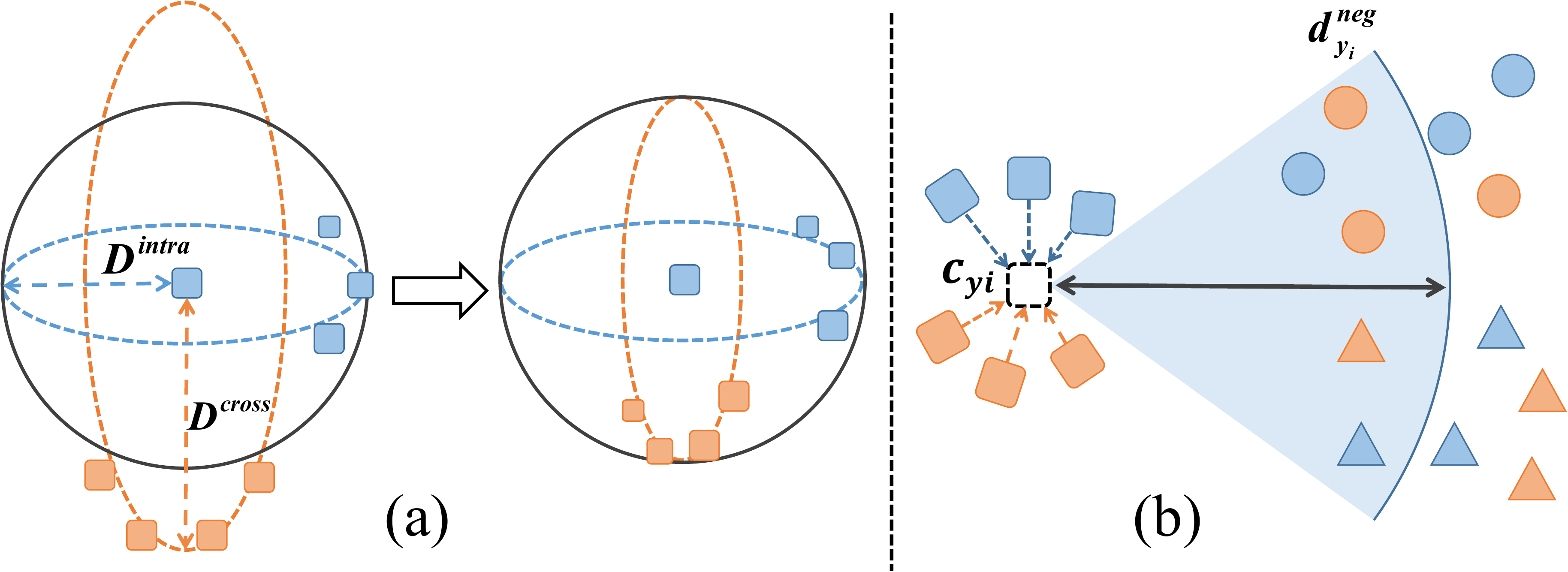}
\caption{Geometric illustrations of the (a) Modality-Shared Enhancement Loss and (b) Discriminative Center Loss. }
\label{fig:msel_dcl}
\end{figure}

In Eq.~\ref{eqmsel}, $L_{MSEL}$ penalizes the difference between $\mathrm{D}^{intra}$ and $\mathrm{D}^{cross}$.
When discriminative features that appear only within one modality, then the difference between $\mathrm{D}^{intra}$ and $\mathrm{D}^{cross}$ will increase, and such anomalies will be captured by $L_{MSEL}$.
During the bi-direction optimization process of $\mathrm{D}^{intra}$ and $\mathrm{D}^{cross}$, the unreliable features that appear in only one modality will be suppressed, while the more reliable features that appear in both modalities will be enhanced as shown in Fig.~\ref{fig:msel} (b).
%
%
Fig.~\ref{fig:msel_dcl} (a) shows the geometric illustrations of the MSEL.
It encourages feature embeddings subject to a spherical distribution.
\subsection{Discriminative Center Loss}
Similar to visible-visible person ReID, the same person may suffer large intra-class differences due to typical variations in pose, point of view, illumination, etc.
They greatly increase the difficulty of feature alignment between different modalities.
%
%
To address this problem, we propose a Discriminative Center Loss (DCL) to exploit the example relationships between center instances and enhance the discriminative power of reliable modality-shared features.

Firstly, to obtain the robust representation of each identity, we compute the feature center under the two modalities by:
\begin{footnotesize}
\begin{equation}
	c_{y_{i}}=\frac{1}{2K}\left(\sum_{j=1}^{\mathrm{K}} f_{j}^{vis}+\sum_{k=1}^{\mathrm{K}} f_{k}^{ir}\right).
\end{equation}
\end{footnotesize}
Here, $c_{y_{i}}$ denotes the feature center of the $y_{i}^{th}$ identity. Then we calculate the average distance of $c_{y_{i}}$ to all other negative samples as a dynamic margin, which can be denoted as:
\begin{footnotesize}
\begin{equation}
	d_{y_{i}}^{neg} =\frac{1}{2K(P-1)}\sum_{\forall y_{j} \neq y_{i}}\left\|f_{j}-c_{y_{i}}\right\|_{2}.
\end{equation}
\end{footnotesize}
Finally, the $L_{DCL}$ is defined as:
\begin{footnotesize}
\begin{equation}\label{eqdcl}
	L_{DCL}= \frac{\sum_{i=1}\limits^{P} \mathop{mean}\limits_{y_j = y_i} \left\|f_{j}-c_{y_{i}}\right\|_{2}}
	{\sum_{i=1}\limits^{P} \mathop{mean}\limits_{
	\left\|f_{k}-c_{y_{i}}\right\|_{2}<d_{y_{i}}^{neg}
	\atop y_k\neq y_i } \left\|f_{k}-c_{y_{i}}\right\|_{2}
    }.
\end{equation}
\end{footnotesize}
By minimizing Eq.~\ref{eqdcl}, the intra-class compactness and inter-class separability will be improved.
Fig.~\ref{fig:msel_dcl} (b) shows the geometric illustrations of the DCL.
The utilization of $L_{DCL}$ has two main advantages:
1) It can utilize modality-specific features and capture more potential relationships than the center-center solution.
2) The dynamic sampling through $d_{y_{i}}^{neg}$ can effectively focus on relatively difficult examples.
The effectiveness will be verified by experiments.
\subsection{Overall Objective Function}
For model training, we adopt a hybrid loss function for our progressive learning framework.
At the first stage, we utilize the identity loss ${L}_{ID}$~\cite{zheng2017discriminatively} and $L_{Intra}$ to learn modality-independent features:
\begin{footnotesize}
\begin{equation}\label{eq}
L_{1}=L_{Intra}+L_{ID}.
\end{equation}
\end{footnotesize}
At the second stage, we further extract the reliable modality-shared features with $L_{MSEL}$ and enhance the discrimination with $L_{DCL}$.
The loss function can be defined as:
\begin{footnotesize}
\begin{equation}\label{eq}
L_{2}=L_{Global}+\lambda_{1}L_{MSEL}+\lambda_{2}L_{DCL}.
\end{equation}
\end{footnotesize}
Here, the parameters $\lambda_{1}$ and $\lambda_{2}$ are used to balance the terms of $L_{MSEL}$ and $L_{DCL}$, respectively.
\section{Experiments}
\subsection{Experimental Setting}
\subsubsection{Datasets.}
In this work, we follow previous methods and conduct experiments on two public VI-ReID datasets.

\textbf{SYSU-MM01}~\cite{wu2017rgb} has a total of 286,628 visible images and 15,792 infrared images with 491 different person identities.
The training set contains 22,258 visible images and 11,909 infrared images of 395 persons, and the testing set contains images of another 96 different identities.
3803 infrared images are used as the query set, and from the other visible images, 301 images are randomly selected as the gallery set.
Besides, there are two search modes.
The \emph{all-search mode} uses all images for testing, while the \emph{indoor-search mode} only uses the indoor images.

\textbf{RegDB}~\cite{nguyen2017person} contains a total of 412 different person identities.
For each person, 10 visible images and 10 infrared images are captured.
We follow the evaluation protocol in~\cite{ye2018hierarchical} and randomly select all images of 206 identities for training and the remaining 206 identities for testing.
To obtain the stable results, we randomly divide this dataset ten times for independent training and testing.
\subsubsection{Evaluation metrics.}
We use Cumulative Matching Characteristics (CMC), Mean Average Precision (mAP), and Mean Inverse Negative Penalty (mINP)~\cite{ye2021deep} as our main evaluation metrics.
\subsubsection{Implementation details.}
Our proposed method is implemented with the Huawei-Mindspore toolbox and one NVIDIA RTX3090 GPU.
We adopt the ViT-B/16~\cite{dosovitskiy2020image} pre-trained on ImageNet~\cite{deng2009imagenet} as our backbone and set the overlap stride to 12 to balance speed and performance.
All person images are resized to 256$\times$128 with horizontal flipping and random erasing for data augmentation.
For infrared images, color jitter and gaussian blur are additionally applied.
The batch size is set to 64, containing a total of 8 different identities.
For each identity, 4 visible images and 4 infrared images are sampled.
We adopt AdamW optimizer with a cosine annealing learning rate scheduler for training.
The basic learning rate is set to $3e^{-4}$ and weight decay is set to $1e^{-4}$.
We train 24 epochs for the SYSU-MM01 and 36 epochs for the RegDB.
For both datasets, the epoch $t$ of the first stage is set to 6, the trade-off parameters $\lambda_{1}$ and $\lambda_{2}$ are set to 0.5, and the margin parameter $m$ is set to 0.1.
The 768-dimensional features after the BN layer are used for testing.
%
\subsection{Ablation Studies}
In this subsection, we conduct experiments to verify the effects of different modules on the SYSU-MM01 dataset under the all-search mode.
\begin{table}[h!]\small
\centering
\caption{Effects of different image modalities.}
\label{pl1}
\renewcommand\arraystretch{1.1}
\setlength{\tabcolsep}{1.5mm}
\begin{tabular}{l |l c c c c c}  %
		\hline
		~~Transition Schemes & Rank-1 & Rank-10 & mAP & mINP \cr
		\hline
		Baseline (RGB) & 52.51 & 88.21 & 51.30 & 38.51 \cr
		Baseline (Gray) & 56.51 & 91.76  & 54.22 & 39.75 \cr
		Baseline (RGB-Gray) & 57.24 & 91.87  & 54.95 & 40.25 \cr
		Baseline (Gray-RGB) & 59.07 & 92.53  & 56.86 & 42.81 \cr
		\hline
\end{tabular}
\vspace{-4mm}
\end{table}
\subsubsection{Effectiveness of the progressive learning strategy.}
We evaluate the effectiveness in terms of both image modality and loss function.
%
%
For the image modality, the comparison results are shown in Tab.~\ref{pl1}.
``Baseline (RGB)'' and ``Baseline (Gray)'' indicates the baseline model trained with the RGB-IR modality and Grayscale-IR modality, respectively.
``Baseline (Gray-RGB)'' means that the first $t$ epochs of training uses the Grayscale-IR modality and the rest uses the RGB-IR modality.
From the results, one can see that directly using the RGB-IR modality and Grayscale-IR modality shows inferior performances.
With a progressive learning strategy, the model can show better results, indicating the effectiveness of reducing the modality differences.
Our proposed strategy significantly improves the ``Baseline (RGB)'' model by 5.16\% Rank-1, 5.04\% mAP, and 5.74\% mINP.

Besides, based on the ``Baseline (Gray-RGB)'' model, we replace the hard triplet loss with WRT \cite{ye2021deep}, HCT \cite{liu2020parameter} and our proposed PHT.
The comparison results are shown in Tab. \ref{pl2}.
These results further prove the effectiveness of our progressive learning strategy.
\begin{table}[h]\small
\centering
\caption{Effects of progressive learning losses.}
\label{pl2}
\renewcommand\arraystretch{1.1}
\setlength{\tabcolsep}{1.5mm}
\begin{tabular}{l |l c c c c c}  %
		\hline
		~~~~~~~Methods & Rank-1 & Rank-10 & mAP & mINP \cr
		\hline
		HardTri ($m = 0.1$) & 59.07 & 92.53  & 56.86 & 42.81 \cr
		WRT & 54.90 & 92.33  & 54.74 & 42.30 \cr
		HCT ($m = 0.3$) & 59.51 & 92.38  & 56.68 & 42.05 \cr
		PHT ($m = 0.1$) & 61.67 & 93.02  & 59.26 & 45.49 \cr
		\hline
\end{tabular}
\end{table}
\subsubsection{Effects with MSEL and DCL.}
Table \ref{dcl} shows the comparison results of different settings with MSEL and DCL.
``Base(PL)'' only adopts the progressive learning strategy.
``MSEL (Cosine)'' and ``MSEL (Euclid)'' mean that the model uses the cosine distance and Euclidean distance, respectively.
%
%
The experimental results show that the model with MSEL consistently improves the performance.
The ``MSEL (Euclid)'' model brings the best results with 3.44\% Rank-1, 2.21\% mAP, and 1.91\% mINP improvement compared to ``Base (PL)''.
\begin{table}[h!]\small
\centering
\caption{Comparison results with MSEL and DCL.}
\label{dcl}
\renewcommand\arraystretch{1.1}
\setlength{\tabcolsep}{1.1mm}
\begin{tabular}{l |l c c c c c}  %
		\hline
		~~~~~Methods & Rank-1 & Rank-10 & mAP & mINP \cr
		\hline
		Base (PL) & 61.67 & 93.02  & 59.26 & 45.49 \cr \hline
		+MSEL (Cosine)&64.19 & 93.45   & 60.67 & 46.15 \cr
		+MSEL (Euclid)&65.11 & 93.81 & 61.47 & 47.40 \cr \hline
		+DCL (Hard) & 63.22 & 94.09  & 62.13 & 49.97 \cr
		+DCL (All) & 62.86 & 94.15  & 61.32 & 48.73  \cr
		+DCL (Dyn) & 64.05 & 94.71  & 62.76 & 50.38 \cr \hline
		+MSEL (Euclid)+DCL (Dyn) & 67.53 & 95.36  & 64.98 & 51.86 \cr \hline
\end{tabular}
\end{table}

As for the DCL, ``DCL (Hard)'' means only selecting the closest negative sample for each identity center.
``DCL (All)'' means selecting all negative samples for each identity center, and ``DCL (Dyn)'' means dynamically selecting negative samples based on Eq.9.
As shown in Tab. \ref{dcl}, the $L_{DCL}$ can bring a consistent improvement.
Our dynamic selection shows much better relative results with 2.38\% Rank-1, 3.50\% mAP, and 4.89\% mINP when compared with ``Base (PL)''.
Combined with MSEL, the model can bring a further 2.42\% Rank-1, 3.51\% mAP and 4.46\% mINP improvement.
These results fully demonstrate the effectiveness of our MSEL and DCL.
\begin{table}[t!]\small
\centering
\caption{Comparison results with CNN-based backbones.}
\label{tab_agw}
\renewcommand\arraystretch{1.1}
\setlength{\tabcolsep}{1.5mm}
\begin{tabular}{l |l c c c c c}
		\hline
		~~~~~Methods & Rank-1 & Rank-10 & mAP & mINP \cr
		\hline
		AGW & 58.19 & 91.22  & 56.50 & 43.52 \cr
		AGW+MSEL & 62.16 & 92.52  & 59.66 & 46.38 \cr
		AGW+MSEL+PL & 65.97 & 94.79  & 62.15 & 47.30 \cr
		AGW+MSEL+PL+DCL & 67.09 & 94.56  & 64.25 & 50.89 \cr
		\hline
		Ours & 67.53 & 95.36  & 64.98 & 51.86 \cr
		\hline
\end{tabular}
\vspace{-4mm}
\end{table}
\subsubsection{Effects of CNN backbones.}
To further study the effectiveness and generalization of our proposed methods, we also carry out experiments with CNN-based frameworks.
As a typical example, we take the outstanding AGW method with random erasing~\cite{ye2021deep}.
%
%
The experimental results are listed in Tab. \ref{tab_agw}.
%
%
%
The results show that by adding MSEL, the model delivers a performance gain of 3.97\% Rank-1, 3.16\% mAP, and 2.86\% mINP.
The results indicate that $L_{MSEL}$ can also be compatible with variants of different triples.
With the full modules (``MSEL+PL+DCL''), the model can achieve a performance gain of 8.90\% Rank-1, 7.75\% mAP, and 8.34\% mINP.
These facts clearly demonstrate the generalization of our proposed methods on CNN-based frameworks.
However, our Transformer-based framework shows better results than CNN-based ones, as shown in the last row.
\subsubsection{Trade-off parameters.}
\begin{figure}[ht]
\centering
\includegraphics[width=0.6\columnwidth, trim=0 0 10 10]{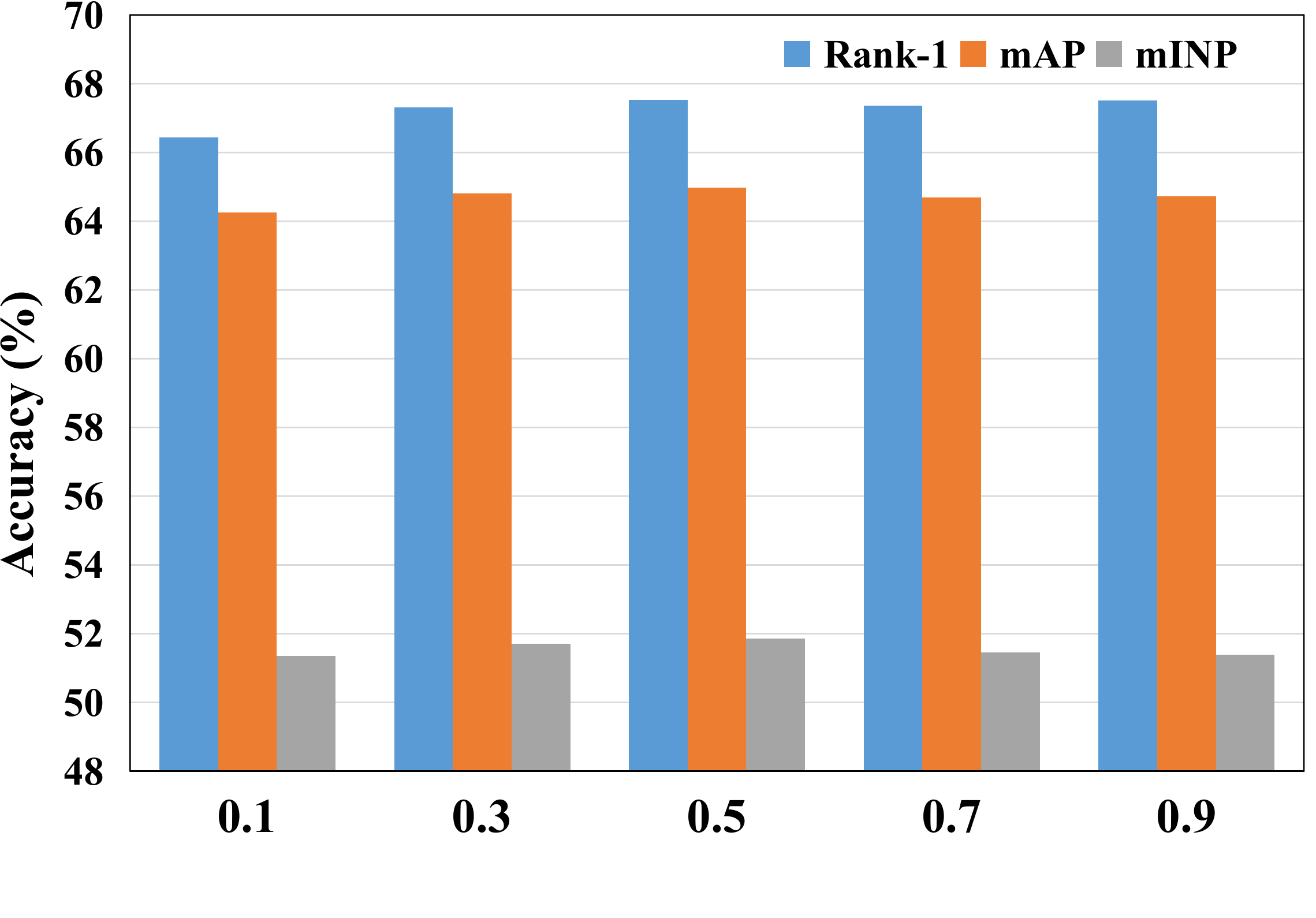}\\
\includegraphics[width=0.6\columnwidth, trim=0 0 10 10]{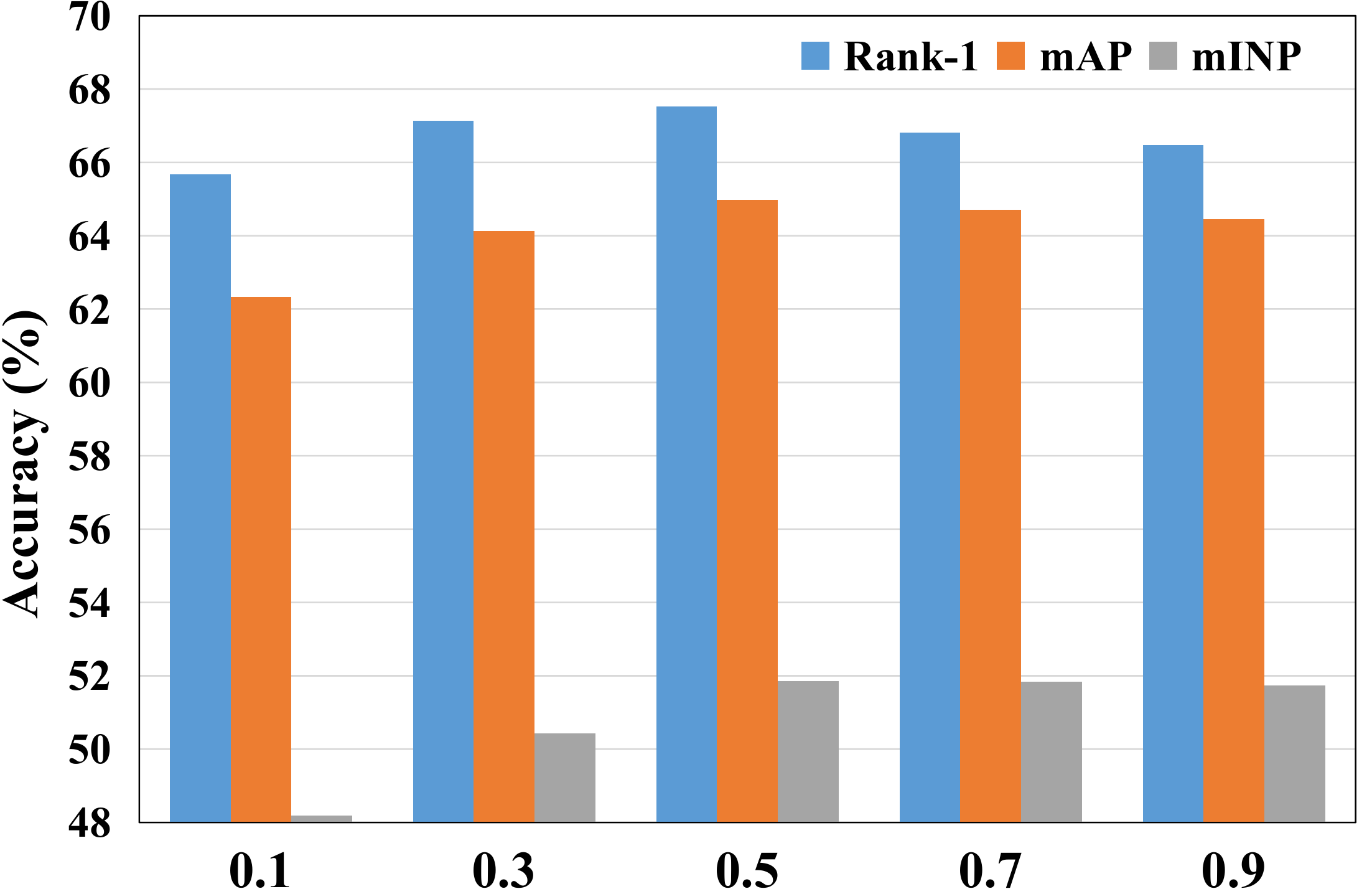}
\caption{Performance effects of trade-off parameters $\lambda_{1}$ and $\lambda_{2}$.
In the top sub-figure, $\lambda_{2}=0.5$, $\lambda_{1}\in$ [0.1, 0.9] and in the bottom sub-figure, $\lambda_{1}= 0.5$, $\lambda_{2}\in$ [0.1, 0.9].}
\label{par}
\end{figure}

We conduct additional experiments to evaluate the effect of trade-off parameters $\lambda_{1}$ and $\lambda_{2}$.
As shown in Fig. \ref{par}, $L_{MSEL}$ is not sensitive to the parameter settings, while $L_{DCL}$ is stable within a certain range.
\subsubsection{Visualization analysis.}
\begin{figure}[h!]
	\centering
	\includegraphics[width=1.0\columnwidth]{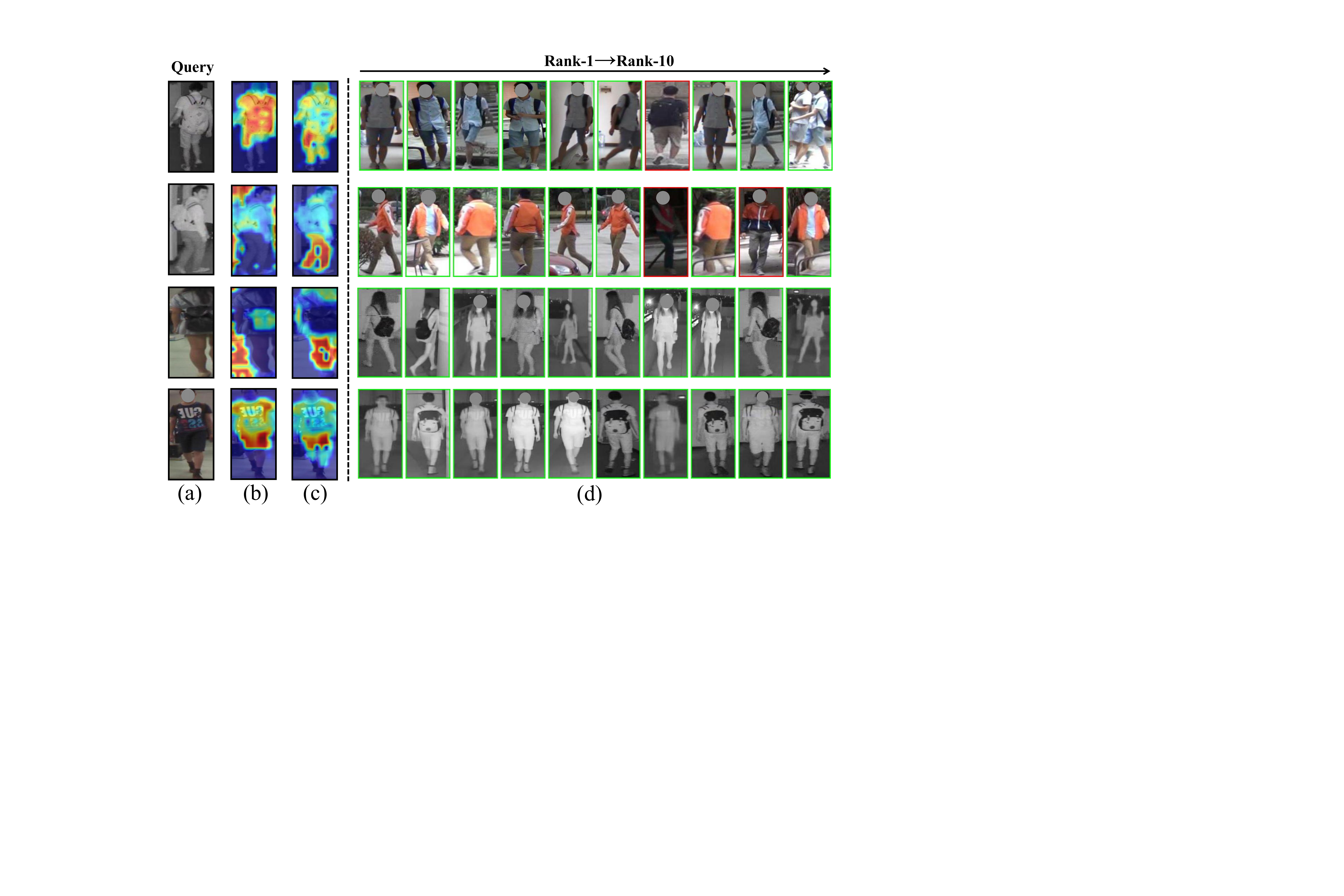}
	\caption{Attention maps and retrieval results. (a) Query images. (b) Attention maps w/o MSEL. (c) Attention maps w/ MSEL. (d) Top 10 retrieval results.}
	\label{vis}
\end{figure}

To analyse the visual effect of our proposed model, we present some typical visual examples.
As shown in Fig.~\ref{vis}, we use the Grad-CAM~\cite{selvaraju2017grad} to generate attention maps of query images with our models.
Besides, the top 10 retrieval results are also provided in Fig.~\ref{vis} (d).
One can observe that with our MSEL, the model focuses on more discriminative regions and extracts decent modality-shared features.
Thus, the model can effectively deal with complex scenarios, as shown in Fig.~\ref{fig:challenges}.
\begin{figure}[t!]
	\centering
	\includegraphics[width=1.0\columnwidth]{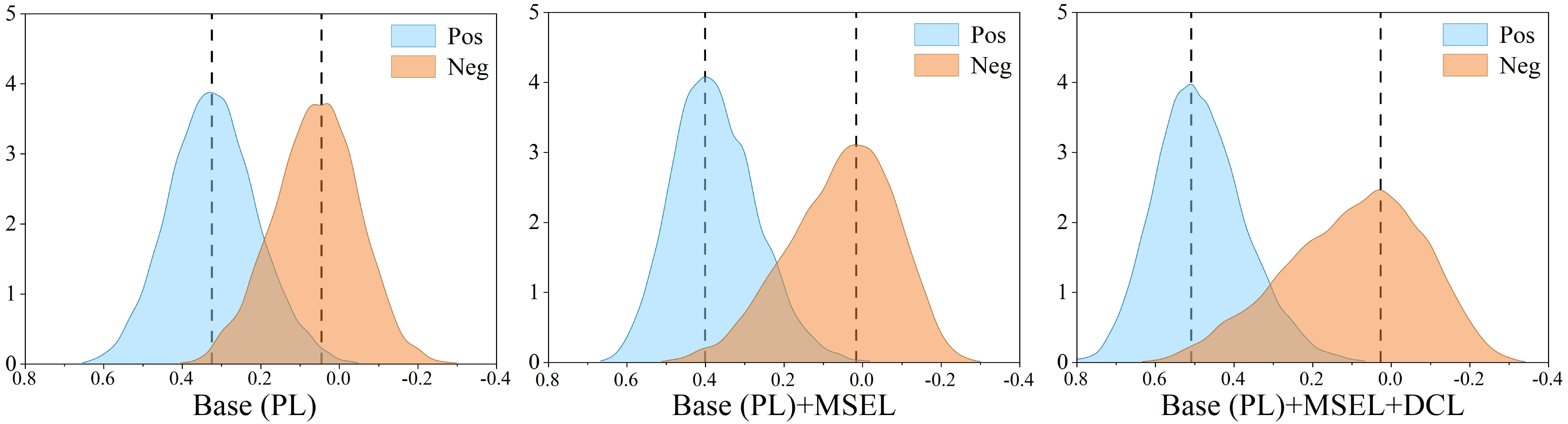}
	\caption{The cosine similarity distribution of positive and negative matching pairs from the testing set.}
	\label{distbut}
\vspace{-4mm}
\end{figure}

In addition, we randomly sample 10,000 positive and negative matching pairs from the testing set and visualize their cosine similarity distribution, as shown in Fig. \ref{distbut}.
One can observe that the similarity of positive cross-modality matching pairs increases by introducing MSEL, which indicates that MSEL enhances the utilization of reliable modality-invariant features.
By further introducing DCL, the range of cosine similarity between negative and positive pairs is significantly expanded.
%
%
The visualization shows the ability of $L_{DCL}$ to explore potential information and effectively improve the discrimination of feature embeddings.
\begin{table*}[t]
	\centering
	\caption{Comparisons with state-of-the-art methods under all-search and indoor-search modes on the SYSU-MM01 dataset.}
	\label{sota_sysu}
	\renewcommand\arraystretch{1.1}
	\setlength{\tabcolsep}{1.5mm}
	\begin{tabular}{l|c|c|c|c|c||c|c|c|c|c}
		\hline
		\multicolumn{1}{c|}{\emph{Methods}}  &
		\multicolumn{5}{c||}{\emph{All search}} & \multicolumn{5}{c}{\emph{Indoor search}} \\
		&   $r = 1$  & $r = 10$ & $r = 20$ & mAP & mINP  &  $r = 1$  & $r = 10$ & $r = 20$ & mAP & mINP     \\ \hline
		Zero-Pad \cite{wu2017rgb} & 14.80 & 54.12 & 71.33 & 15.95 & - & 20.58 & 68.38 & 85.79 & 26.92 &-  \\
		Hi-CMD \cite{choi2020hi} & 34.94 & 77.58 & - & 35.94 & -  &  - & - & - & - & - \\
		CMSP \cite{wu2020rgb} & 43.56  & 86.25 & - & 44.98 &- & 48.62  & 89.50 & -  & 57.50 &- \\
		expAT Loss \cite{ye2020bi} & 38.57 & 76.64 & 86.39 & 38.61 & - & - & - & -& - & - \\
		AGW \cite{ye2021deep} & 47.50 & 84.39 & 92.14 & 47.65 & 35.30 & 54.17 & 91.14 & 95.98 & 62.97 & 59.23 \\
		HAT \cite{ye2020visible} & 55.29 & 92.14 & 97.36 & 53.89 & - & 62.10 & 95.75 & 99.20& 69.37 & - \\
		LBA \cite{park2021learning}  & 55.41 & 91.12 & - & 54.14 & - & 58.46 & 94.13 & - & 66.33 & - \\
		NFS \cite{chen2021neural}  & 56.91 & 91.34 & 96.52 & 55.45 & - & 62.79 & 96.53 & 99.07 & 69.79 & - \\
		MSO \cite{gao2021mso} & 58.70 & 92.06 & 97.20 & 56.42 & 42.04 & 63.09 & 96.61 & -& 70.31 & - \\
		CM-NAS \cite{fu2021cm} & 61.99 & 92.87 & 97.25 & 60.02 & - & 67.01 & \textbf{97.02} & \textbf{99.32} & 72.95 & - \\
		MID \cite{huang2022modality} & 60.27 & 92.90 & - & 59.40 & - & 64.86 & 96.12 & - & 70.12 & - \\
		SPOT \cite{chen2022structure} & 65.34 & 92.73 & 97.04 & 62.25 & 48.86 & 69.42 & 96.22 & 99.12 & 74.63 & 70.48 \\
		FMCNet \cite{zhang2022fmcnet} & 66.34 & - & - & 62.51 & - & 68.15 & - & - & 74.09 & - \\ \hline
		\textbf{PMT (Ours)}  & \textbf{67.53}  & \textbf{95.36} & \textbf{98.64} & \textbf{64.98} & \textbf{51.86} & \textbf{71.66} & 96.73 & 99.25 & \textbf{76.52} & \textbf{72.74} \\
		\hline
	\end{tabular}
\end{table*}
\begin{table}[h!]\small
	\centering
	\caption{Comparison results under Visible-Thermal and
		Thermal-Visible modes on the RegDB dataset.}
	\label{sota_regdb}
	\renewcommand\arraystretch{1.02}
	\setlength{\tabcolsep}{1.2mm}
	\begin{tabular}{l|c|c|c|c c}
		\hline
		\multicolumn{1}{c|}{\emph{Methods}} & \multicolumn{2}{c|}{\emph{V to T}} & \multicolumn{3}{c}{\emph{T to V}} \\
		&  $r = 1$  & mAP &  $r = 1$   & mAP  \\
		\hline
		DDAG \cite{ye2020dynamic}  & 69.34   & 63.46  &  68.06   & 61.80  \\
		expAT Loss \cite{ye2020bi}  & 66.48   & 67.31  & 67.45  & 66.51 \\
		AGW \cite{ye2021deep}  & 70.05  & 66.37  & 70.49   & 65.90  \\
		HAT \cite{ye2020visible}  & 71.83   & 67.56 & 70.02   & 66.30  \\
		MSO \cite{gao2021mso}  & 73.6   & 66.9  & 74.6  & 67.5 & \\
		LBA \cite{park2021learning}  & 74.17   & 67.64  & 72.43   & 65.46  \\
		NFS \cite{chen2021neural}  & 80.54   & 72.10  & 77.95   & 69.79  \\
		MCLNet \cite{hao2021cross}  & 80.31  & 73.07  & 75.93  & 69.49  \\
		SPOT \cite{chen2022structure}  & 80.35   & 72.46  & 79.37  & 72.26 \\ \hline
		\textbf{PMT (Ours)} & \textbf{84.83}  &  \textbf{76.55}  & \textbf{84.16}  & \textbf{75.13}  \\
		\hline
	\end{tabular}
\end{table}
\subsection{Comparison with State-of-the-Arts}
In this subsection, we compare our proposed PMT with other state-of-the-art methods on SYSU-MM01 and RegDB.
\subsubsection{SYSU-MM01:}
Tab. \ref{sota_sysu} shows the comparison results on the SYSU-MM01 dataset.
One can observe that our proposed method outperforms other weight-shared methods [expAT \cite{ye2020bi}, HAT \cite{ye2020visible}] by at least 12.24\% in Rank-1 and 11.09\% in mAP under the all-search mode.
Moreover, compared with dual-stream-based methods [LBA \cite{park2021learning}, NFS \cite{chen2021neural}, SPOT \cite{chen2022structure}], our proposed method also has substantial performance advantages.
Under the indoor-search mode, our proposed method shows much better results in terms of Rank-1, mAP and mINP.
In terms of Rank-10 and Rank-20, CM-NAS \cite{fu2021cm} shows best results.
The main reason may be that the searched network by CM-NAS is more helpful for global discrimination.
However, our proposed method delivers very comparable results.
All the above results fully demonstrate that our proposed PMT can effectively reduce the large modality gap and utilize more reliable modality-shared features.
\subsubsection{RegDB:}
In Tab. \ref{sota_regdb}, we report the comparison results on the RegDB dataset.
The results show that our proposed method achieves excellent performances in both Visible to Thermal ($V to T$) and Thermal to Visible ($T to V$) modes.
More specifically, our proposed method achieves an expressive performance of 84.83\% Rank-1 and 76.55\% mAP under the $V to T$ mode, showing a 4\% performance gain than other best methods.
For the more challenging $T to V$ mode, our method also shows great performance advantages.
These results also indicate that our proposed method is more robust against different datasets and query patterns.
\section{Conclusion}
In this paper, we propose a novel deep learning-based framework named PMT, which effectively improves the performance of VI-ReID by fully exploring reliable modality-invariant features.
With gray-scale images as an auxiliary modality, our framework mitigates the large gap between RGB-IR modalities through a progressive learning strategy.
Meanwhile, our proposed MSEL and DCL can effectively extract more reliable and discriminative features, bringing stronger performance and robustness.
Moreover, the proposed methods have a good generalization.
By applying our methods to CNN-based backbones, they can also bring significant performance improvements.
Experimental results on two public VI-ReID benchmarks verify the effectiveness of our proposed framework.
In the future, we will explore more effective Transformer structures to further improve the feature representation ability.
\section{Acknowledgments}
This work was supported in part by the National Natural Science Foundation of China (NSFC) (No. 91538201, 62101092), the CAAI-Huawei MindSpore Open Fund under Grant CAAIXSJLJJ-2021-067A, and the Fundamental Research Funds for the Central Universities (No. DUT20RC(3)083).
\bibliography{aaai23}

\begin{thebibliography}{37}
\providecommand{\natexlab}[1]{#1}

\bibitem[{Chen et~al.(2022)Chen, Ye, Qi, Wu, Jiang, and
  Lin}]{chen2022structure}
Chen, C.; Ye, M.; Qi, M.; Wu, J.; Jiang, J.; and Lin, C.-W. 2022.
\newblock Structure-Aware Positional Transformer for Visible-Infrared Person
  Re-Identification.
\newblock \emph{IEEE Transactions on Image Processing}, 31: 2352--2364.

\bibitem[{Chen et~al.(2021{\natexlab{a}})Chen, Xu, Xu, and Gao}]{chen2021oh}
Chen, X.; Xu, J.; Xu, J.; and Gao, S. 2021{\natexlab{a}}.
\newblock OH-Former: Omni-Relational High-Order Transformer for Person
  Re-Identification.
\newblock \emph{arXiv:2109.11159}.

\bibitem[{Chen et~al.(2021{\natexlab{b}})Chen, Wan, Li, Jing, and
  Sun}]{chen2021neural}
Chen, Y.; Wan, L.; Li, Z.; Jing, Q.; and Sun, Z. 2021{\natexlab{b}}.
\newblock Neural feature search for rgb-infrared person re-identification.
\newblock In \emph{Proceedings of the IEEE/CVF Conference on Computer Vision
  and Pattern Recognition}, 587--597.

\bibitem[{Choi et~al.(2020)Choi, Lee, Kim, Kim, and Kim}]{choi2020hi}
Choi, S.; Lee, S.; Kim, Y.; Kim, T.; and Kim, C. 2020.
\newblock Hi-CMD: Hierarchical cross-modality disentanglement for
  visible-infrared person re-identification.
\newblock In \emph{Proceedings of the IEEE/CVF Conference on Computer Vision
  and Pattern Recognition}, 10257--10266.

\bibitem[{Dai et~al.(2018)Dai, Ji, Wang, Wu, and Huang}]{dai2018cross}
Dai, P.; Ji, R.; Wang, H.; Wu, Q.; and Huang, Y. 2018.
\newblock Cross-modality person re-identification with generative adversarial
  training.
\newblock In \emph{International Joint Conference on Artificial Intelligence},
  volume~1, 6.

\bibitem[{Deng et~al.(2009)Deng, Dong, Socher, Li, Li, and
  Fei-Fei}]{deng2009imagenet}
Deng, J.; Dong, W.; Socher, R.; Li, L.-J.; Li, K.; and Fei-Fei, L. 2009.
\newblock Imagenet: A large-scale hierarchical image database.
\newblock In \emph{IEEE Conference on Computer Vision and Pattern Recognition},
  248--255. IEEE.

\bibitem[{Dosovitskiy et~al.(2020)Dosovitskiy, Beyer, Kolesnikov, Weissenborn,
  Zhai, Unterthiner, Dehghani, Minderer, Heigold, Gelly
  et~al.}]{dosovitskiy2020image}
Dosovitskiy, A.; Beyer, L.; Kolesnikov, A.; Weissenborn, D.; Zhai, X.;
  Unterthiner, T.; Dehghani, M.; Minderer, M.; Heigold, G.; Gelly, S.; et~al.
  2020.
\newblock An Image is Worth 16x16 Words: Transformers for Image Recognition at
  Scale.
\newblock In \emph{International Conference on Learning Representations}.

\bibitem[{Fu et~al.(2021)Fu, Hu, Wu, Shi, Mei, and He}]{fu2021cm}
Fu, C.; Hu, Y.; Wu, X.; Shi, H.; Mei, T.; and He, R. 2021.
\newblock CM-NAS: Cross-modality neural architecture search for
  visible-infrared person re-identification.
\newblock In \emph{Proceedings of the IEEE/CVF International Conference on
  Computer Vision}, 11823--11832.

\bibitem[{Gao et~al.(2021)Gao, Liang, Jin, Gu, Liu, Li, and Lang}]{gao2021mso}
Gao, Y.; Liang, T.; Jin, Y.; Gu, X.; Liu, W.; Li, Y.; and Lang, C. 2021.
\newblock MSO: Multi-feature space joint optimization network for rgb-infrared
  person re-identification.
\newblock In \emph{Proceedings of the 29th ACM International Conference on
  Multimedia}, 5257--5265.

\bibitem[{Hao et~al.(2021)Hao, Zhao, Ye, and Shen}]{hao2021cross}
Hao, X.; Zhao, S.; Ye, M.; and Shen, J. 2021.
\newblock Cross-modality person re-identification via modality confusion and
  center aggregation.
\newblock In \emph{Proceedings of the IEEE/CVF International Conference on
  Computer Vision}, 16403--16412.

\bibitem[{He et~al.(2021)He, Luo, Wang, Wang, Li, and Jiang}]{he2021transreid}
He, S.; Luo, H.; Wang, P.; Wang, F.; Li, H.; and Jiang, W. 2021.
\newblock Transreid: Transformer-based object re-identification.
\newblock In \emph{Proceedings of the IEEE/CVF International Conference on
  Computer Vision}, 15013--15022.

\bibitem[{Huang et~al.(2022)Huang, Liu, Li, Zheng, and Zha}]{huang2022modality}
Huang, Z.; Liu, J.; Li, L.; Zheng, K.; and Zha, Z.-J. 2022.
\newblock Modality-Adaptive Mixup and Invariant Decomposition for RGB-Infrared
  Person Re-Identification.
\newblock \emph{arXiv preprint arXiv:2203.01735}.

\bibitem[{Lai, Chai, and Wei(2021)}]{lai2021transformer}
Lai, S.; Chai, Z.; and Wei, X. 2021.
\newblock Transformer Meets Part Model: Adaptive Part Division for Person
  Re-Identification.
\newblock In \emph{Proceedings of the IEEE/CVF International Conference on
  Computer Vision}, 4150--4157.

\bibitem[{Li et~al.(2020)Li, Wei, Hong, and Gong}]{li2020infrared}
Li, D.; Wei, X.; Hong, X.; and Gong, Y. 2020.
\newblock Infrared-visible cross-modal person re-identification with an x
  modality.
\newblock In \emph{Proceedings of the AAAI Conference on Artificial
  Intelligence}, volume~34, 4610--4617.

\bibitem[{Liang et~al.(2021)Liang, Jin, Gao, Liu, Feng, Wang, and
  Li}]{liang2021cmtr}
Liang, T.; Jin, Y.; Gao, Y.; Liu, W.; Feng, S.; Wang, T.; and Li, Y. 2021.
\newblock CMTR: Cross-modality Transformer for Visible-infrared Person
  Re-identification.
\newblock \emph{arXiv preprint arXiv:2110.08994}.

\bibitem[{Liu, Tan, and Zhou(2020)}]{liu2020parameter}
Liu, H.; Tan, X.; and Zhou, X. 2020.
\newblock Parameter sharing exploration and hetero-center triplet loss for
  visible-thermal person re-identification.
\newblock \emph{IEEE Transactions on Multimedia}, 23: 4414--4425.

\bibitem[{Lu et~al.(2020)Lu, Wu, Liu, Zhang, Li, Chu, and Yu}]{lu2020cross}
Lu, Y.; Wu, Y.; Liu, B.; Zhang, T.; Li, B.; Chu, Q.; and Yu, N. 2020.
\newblock Cross-modality person re-identification with shared-specific feature
  transfer.
\newblock In \emph{Proceedings of the IEEE/CVF Conference on Computer Vision
  and Pattern Recognition}, 13379--13389.

\bibitem[{Ma, Zhao, and Li(2021)}]{ma2021pose}
Ma, Z.; Zhao, Y.; and Li, J. 2021.
\newblock Pose-guided inter-and intra-part relational transformer for occluded
  person re-identification.
\newblock In \emph{Proceedings of the 29th ACM International Conference on
  Multimedia}, 1487--1496.

\bibitem[{Nguyen et~al.(2017)Nguyen, Hong, Kim, and Park}]{nguyen2017person}
Nguyen, D.~T.; Hong, H.~G.; Kim, K.~W.; and Park, K.~R. 2017.
\newblock Person recognition system based on a combination of body images from
  visible light and thermal cameras.
\newblock \emph{Sensors}, 17(3): 605.

\bibitem[{Park et~al.(2021)Park, Lee, Lee, and Ham}]{park2021learning}
Park, H.; Lee, S.; Lee, J.; and Ham, B. 2021.
\newblock Learning by aligning: Visible-infrared person re-identification using
  cross-modal correspondences.
\newblock In \emph{Proceedings of the IEEE/CVF International Conference on
  Computer Vision}, 12046--12055.

\bibitem[{Selvaraju et~al.(2017)Selvaraju, Cogswell, Das, Vedantam, Parikh, and
  Batra}]{selvaraju2017grad}
Selvaraju, R.~R.; Cogswell, M.; Das, A.; Vedantam, R.; Parikh, D.; and Batra,
  D. 2017.
\newblock Grad-cam: Visual explanations from deep networks via gradient-based
  localization.
\newblock In \emph{Proceedings of the IEEE International Conference on Computer
  Vision}, 618--626.

\bibitem[{Vaswani et~al.(2017)Vaswani, Shazeer, Parmar, Uszkoreit, Jones,
  Gomez, Kaiser, and Polosukhin}]{vaswani2017attention}
Vaswani, A.; Shazeer, N.; Parmar, N.; Uszkoreit, J.; Jones, L.; Gomez, A.~N.;
  Kaiser, {\L}.; and Polosukhin, I. 2017.
\newblock Attention is all you need.
\newblock \emph{Advances in Neural Information Processing Systems}, 30.

\bibitem[{Wang et~al.(2019)Wang, Zhang, Cheng, Liu, Yang, and
  Hou}]{wang2019rgb}
Wang, G.; Zhang, T.; Cheng, J.; Liu, S.; Yang, Y.; and Hou, Z. 2019.
\newblock RGB-infrared cross-modality person re-identification via joint pixel
  and feature alignment.
\newblock In \emph{Proceedings of the IEEE/CVF International Conference on
  Computer Vision}, 3623--3632.

\bibitem[{Wu et~al.(2020)Wu, Zheng, Gong, and Lai}]{wu2020rgb}
Wu, A.; Zheng, W.-S.; Gong, S.; and Lai, J. 2020.
\newblock Rgb-ir person re-identification by cross-modality similarity
  preservation.
\newblock \emph{International Journal of Computer Vision}, 128(6): 1765--1785.

\bibitem[{Wu et~al.(2017)Wu, Zheng, Yu, Gong, and Lai}]{wu2017rgb}
Wu, A.; Zheng, W.-S.; Yu, H.-X.; Gong, S.; and Lai, J. 2017.
\newblock RGB-infrared cross-modality person re-identification.
\newblock In \emph{Proceedings of the IEEE International Conference on Computer
  Vision}, 5380--5389.

\bibitem[{Ye et~al.(2020{\natexlab{a}})Ye, Liu, Meng, and Li}]{ye2020bi}
Ye, H.; Liu, H.; Meng, F.; and Li, X. 2020{\natexlab{a}}.
\newblock Bi-directional exponential angular triplet loss for RGB-infrared
  person re-identification.
\newblock \emph{IEEE Transactions on Image Processing}, 30: 1583--1595.

\bibitem[{Ye et~al.(2018)Ye, Lan, Li, and Yuen}]{ye2018hierarchical}
Ye, M.; Lan, X.; Li, J.; and Yuen, P. 2018.
\newblock Hierarchical discriminative learning for visible thermal person
  re-identification.
\newblock In \emph{Proceedings of the AAAI Conference on Artificial
  Intelligence}, volume~32.

\bibitem[{Ye et~al.(2020{\natexlab{b}})Ye, Shen, J~Crandall, Shao, and
  Luo}]{ye2020dynamic}
Ye, M.; Shen, J.; J~Crandall, D.; Shao, L.; and Luo, J. 2020{\natexlab{b}}.
\newblock Dynamic dual-attentive aggregation learning for visible-infrared
  person re-identification.
\newblock In \emph{European Conference on Computer Vision}, 229--247. Springer.

\bibitem[{Ye et~al.(2021)Ye, Shen, Lin, Xiang, Shao, and Hoi}]{ye2021deep}
Ye, M.; Shen, J.; Lin, G.; Xiang, T.; Shao, L.; and Hoi, S.~C. 2021.
\newblock Deep learning for person re-identification: A survey and outlook.
\newblock \emph{IEEE Transactions on Pattern Analysis and Machine
  Intelligence}, 44(6): 2872--2893.

\bibitem[{Ye, Shen, and Shao(2020)}]{ye2020visible}
Ye, M.; Shen, J.; and Shao, L. 2020.
\newblock Visible-infrared person re-identification via homogeneous augmented
  tri-modal learning.
\newblock \emph{IEEE Transactions on Information Forensics and Security}, 16:
  728--739.

\bibitem[{Zhang et~al.(2021{\natexlab{a}})Zhang, Zhang, Qi, and
  Lu}]{zhang2021hat}
Zhang, G.; Zhang, P.; Qi, J.; and Lu, H. 2021{\natexlab{a}}.
\newblock Hat: Hierarchical aggregation transformers for person
  re-identification.
\newblock In \emph{Proceedings of the 29th ACM International Conference on
  Multimedia}, 516--525.

\bibitem[{Zhang et~al.(2021{\natexlab{b}})Zhang, Du, Liu, Tu, and
  Shu}]{zhang2021global}
Zhang, L.; Du, G.; Liu, F.; Tu, H.; and Shu, X. 2021{\natexlab{b}}.
\newblock Global-local multiple granularity learning for cross-modality
  visible-infrared person reidentification.
\newblock \emph{IEEE Transactions on Neural Networks and Learning Systems}.

\bibitem[{Zhang et~al.(2022)Zhang, Lai, Liu, Huang, and Han}]{zhang2022fmcnet}
Zhang, Q.; Lai, C.; Liu, J.; Huang, N.; and Han, J. 2022.
\newblock FMCNet: Feature-Level Modality Compensation for Visible-Infrared
  Person Re-Identification.
\newblock In \emph{Proceedings of the IEEE/CVF Conference on Computer Vision
  and Pattern Recognition}, 7349--7358.

\bibitem[{Zheng, Zheng, and Yang(2017)}]{zheng2017discriminatively}
Zheng, Z.; Zheng, L.; and Yang, Y. 2017.
\newblock A discriminatively learned cnn embedding for person reidentification.
\newblock \emph{ACM Transactions on Multimedia Computing, Communications, and
  Applications (TOMM)}, 14(1): 1--20.

\bibitem[{Zhong et~al.(2020)Zhong, Zheng, Kang, Li, and Yang}]{zhong2020random}
Zhong, Z.; Zheng, L.; Kang, G.; Li, S.; and Yang, Y. 2020.
\newblock Random erasing data augmentation.
\newblock In \emph{Proceedings of the AAAI Conference on Artificial
  Intelligence}, volume~34, 13001--13008.

\bibitem[{Zhu et~al.(2021)Zhu, Guo, Zhang, Wang, Huang, Qiao, Liu, Wang, and
  Tang}]{zhu2021aaformer}
Zhu, K.; Guo, H.; Zhang, S.; Wang, Y.; Huang, G.; Qiao, H.; Liu, J.; Wang, J.;
  and Tang, M. 2021.
\newblock Aaformer: Auto-aligned transformer for person re-identification.
\newblock \emph{arXiv preprint arXiv:2104.00921}.

\bibitem[{Zhu et~al.(2020)Zhu, Yang, Wang, Zhao, Hu, and Tao}]{zhu2020hetero}
Zhu, Y.; Yang, Z.; Wang, L.; Zhao, S.; Hu, X.; and Tao, D. 2020.
\newblock Hetero-center loss for cross-modality person re-identification.
\newblock \emph{Neurocomputing}, 386: 97--109.

\end{thebibliography}
\end{document}